\documentclass[11pt,a4paper]{article}
\usepackage[hyperref]{emnlp2020}
\usepackage{times}
\usepackage{latexsym}
\usepackage{graphicx}
\graphicspath{ {./images/} }

\usepackage{amsthm}
\usepackage{amsmath}
\usepackage{amssymb}
\usepackage{color}
\usepackage[skip=10pt]{subcaption}
\usepackage{microtype}
\usepackage{enumitem}

\aclfinalcopy 
\def\aclpaperid{***} 

\title{Inferring symmetry in natural language}

\author{
  Chelsea Tanchip\thanks{\indent Equal contribution.} $^{,\,1}$, Lei Yu\footnotemark[1] $^{,\,2}$, Aotao Xu$^2$, Yang Xu$^{2,\,3,\,4}$ \\ \\
  $^1$ Department of Speech-Language Pathology, University of Toronto, Toronto, Canada \\
  $^2$ Department of Computer Science, University of Toronto, Toronto, Canada\\
  $^3$ Cognitive Science Program, University of Toronto, Toronto, Canada\\
  $^4$ Vector Institute for Artificial Intelligence, Toronto, Canada \\
  \texttt{c.tanchip@mail.utoronto.ca} \\
  \texttt{\{jadeleiyu,a26xu,yangxu\} @cs.toronto.edu}
 }
\date{}

\begin{document}
\maketitle
\begin{abstract}
 
We present a methodological framework for inferring symmetry of verb predicates in natural language. Empirical work on predicate symmetry has taken two main approaches. The feature-based approach focuses on linguistic features pertaining to symmetry. The context-based approach denies the existence of absolute symmetry but instead argues that such inference is context dependent. We develop methods that formalize these approaches and evaluate them against a novel symmetry inference sentence (SIS) dataset comprised of 400 naturalistic usages of literature-informed verbs spanning the spectrum of symmetry-asymmetry. Our results show that a hybrid transfer learning model that integrates linguistic features with contextualized language models most faithfully predicts the empirical data. Our work integrates existing approaches to symmetry in natural language and suggests how symmetry inference can improve systematicity in state-of-the-art language models.
\end{abstract}


\section{Introduction} 

Symmetry helps one make systematic inference about relations in the world and is a fundamental property of natural language (Gleitman, Senghas, Flaherty, Coppola, \& Goldin-Meadow, 2019). A symmetrical predicate describes a reciprocal relation and collective participation between entities. In logical terms, given a symmetrical relation $R$, for all entities $x, y$: $R(x,y) \Longleftrightarrow R(y,x)$. For instance, knowing  \textit{John met Mark} one can systematically infer that \textit{Mark met John}, and vice versa. Here \textit{meet} is perceived as symmetrical, because a meeting is implicitly reciprocal and occurring collectively with both participants. Conversely, \textit{Gab kissed Anna} does not  imply  that \textit{Anna kissed Gab}. Here \textit{kiss} is perceived as  asymmetrical. However, symmetry inference concerns beyond a predicate. In particular, context can make {\it kiss} symmetrical, e.g., \textit{Anna and Gab kissed simultaneously} implies that Anna kissed Gab and Gab kissed Anna. We present a  framework for automated inference of verb symmetry in naturalistic sentences.

Empirical studies from psycholinguistics have taken two main approaches to sentence-level symmetry: 1) a feature-based approach (Gleitman, Gleitman, Miller, \& Ostrin, 1996); and 2) a context-based approach (Tversky \& Gati, 1978). Gleitman and colleagues, after obtaining predicate-level symmetry ratings, had participants assess the degree of discrepancy in meaning between a sentence and its reversed counterpart (where the positions of the entities are switched). The logic behind this approach to symmetry inference can be demonstrated in the pair of sentences, \textit{Gab kissed Anna} and \textit{Anna kissed Gab}, which do not have the same meaning. The difference score for the pair would be high, rendering \textit{kiss} asymmetrical.

{\bf The feature-based approach.} Gleitman and colleagues (\citeyear{Gleitman1996SimilarAS}) found that sentence interpretation heavily depends on its syntactic structure and the lexical-semantic properties of the predicate and entities involved. For example, any predicate can appear symmetrical in a non-directional sentence format (where the entities are placed on one side of the verb, e.g., \textit{Anna and Gab kissed}). Gleitman and colleagues' work suggests that symmetric inference is grounded in linguistic features. However, their findings were based purely on empirical investigation, and no formal approach has been developed to model symmetric inference in language and evaluated comprehensively against data.

The feature-based approach is insufficient to capture all possible real-world relations between entities. As \citet{Gleitman1996SimilarAS} noted, context becomes relevant to determine degree of predicate symmetry such as in the following pair of sentences: \textit{My sister met Meryl Streep} (judged asymmetric) and \textit{John met Mark} (judged symmetric), which indicates that sentences similar in lexical and syntactic features do not always yield the same symmetry judgment.

{\bf The context-based approach.} Focusing on the symmetric predicate {\it similar} instead of verb predicates in their generality, \citet{Tversky1978StudiesOS} elaborated further on the role of context. First they examined  the nature of entities. They deliberately chose entities that are conceptually close in prominence (e.g. \textit{Austria}, \textit{West Germany}) or much different (e.g. \textit{England}, \textit{Jordan}), and found that symmetric inference can depend on one's world knowledge. In a related experiment, they showed that inference involving the predicate {\it similar} can be manipulated with contextual information. For example, {\it Hungary} was judged to be more similar to {\it Austria} than {\it Sweden} or \textit{Norway}, but \textit{Sweden} was judged to be more similar to {\it Austria} than Soviet-aligned {\it Hungary} or Soviet-aligned \textit{Poland}. This approach highlights the need to formalize a contextual approach to symmetry and evaluate how it interacts and fairs with the feature approach.


Our view is that both linguistic features and contextual knowledge matter in symmetry judgment, and integrating the two approaches described should facilitate systematic inference \cite{fodor1987psychosemantics} in models of natural language processing (NLP). We develop a naturalistic sentence dataset for symmetry inference of literature-informed verbs spanning symmetry-asymmetry that is under-represented in existing natural language inference datasets such as SNLI \cite{snli:emnlp2015}. We show that whereas a contextualized language model helps operationalize a context-based approach to symmetry inference, it is critically lacking in learning linguistic features pertaining to symmetry. We propose a hybrid transfer learning model that integrates linguistic features with context and demonstrate its efficacy in improving systematic inference of contextual language models.


\section{Related work}

\subsection{Symmetry in logic vs. empirical tradition} 

In logic, symmetry and reciprocity \cite{Siloni2012ReciprocalVA, Winter2018SymmetricPA} are treated differently, but the difference is often overlooked in empirical tasks. Symmetrical predicates describe a collective event encompassing all entities involved, while reciprocity relates propositions \cite{Gleitman2019TheEO}. In other words, symmetry describes one event and reciprocity describes multiple events occurring with the same action and the same entities but only with roles reversed. To exemplify the difference, take the following sentences: \textit{John and Mary hug} and \textit{John and Mary hug each other}. The first sentence is symmetric and reciprocal, as hugging here is one event with simultaneous reciprocation. The second sentence, however, arguably describes two separate events occurring sequentially: {\it hug(John,Mary)} and then {\it hug(Mary,John)} \cite{Winter2018SymmetricPA}. The difference between symmetry and reciprocity is not syntactically obvious, which is why humans tend to treat the two concepts as the same in sentence-only tasks \cite{Gleitman1996SimilarAS}. Empirical studies have since used visual stimuli to help participants separate symmetry and reciprocity \cite{kruitwagen2017reciprocal, Majid2011TheSO}.
Given these findings, we do not expect human judgment to differentiate symmetry and reciprocity problem from sentence-only stimuli. However, it is instructive to explore how  NLP models, particularly contextualized language models such as BERT \cite{devlin2018bert}, would fare in these cases.

\subsection{Symmetry and systematicity in natural language inference} 

Psycholinguistic research suggests that conceiving symmetry relations relies on essential human capabilities of language understanding. However, few studies have modelled symmetry inference computationally or tested models against empirical data. Symmetry inference can be treated as a special case of recognizing textual entailment (RTE): the pair of input sentences for symmetry problems are typically identical, except that the entities (e.g., subject and object) associated with the target predicate are permuted. Existing studies in semantic inference have constructed NLP systems to predict entailment directionality between simple expressions \cite{bhagat2007ledir}. However, their methods often rely  on human-annotated  features and  fail on more complex examples where contextual dependency is essential for entailment recognition. 

Deep contextualized language models have since been shown to capture rich contextual information in various natural language inference (NLI) tasks, which is a promising starting point for modelling symmetry in natural context \cite{Peters:2018emlo}. However, the interpretability and robustness of these large-scale pre-trained models are yet to be evaluated on symmetry inference. In a series of case studies, Goodwin and colleagues (\citeyear{Goodwin2020ProbingLS}) demonstrated that despite the high overall performance, state-of-the-art NLI systems consistently failed to capture the contribution of certain classes of words or regularities in semantic representation. The inability to generalize systematically  is also observed when training sequence-to-sequence neural models to understand instructions with compositional semantic structures \cite{lake2018generalization}. Our methodological framework for symmetry inference is intimately related to systematicity in NLI. A systematic learner should be able to infer for instance that {\it I kissed her} has a higher degree of asymmetry than {\it We kissed each other}. In a comprehensive set of analyses, we demonstrate that both contextual and linguistic cues are essential for accurate inference about symmetry, and a joint approach helps to improve inference in contextualized language models.

\begin{figure}
\includegraphics[width=0.48\textwidth]{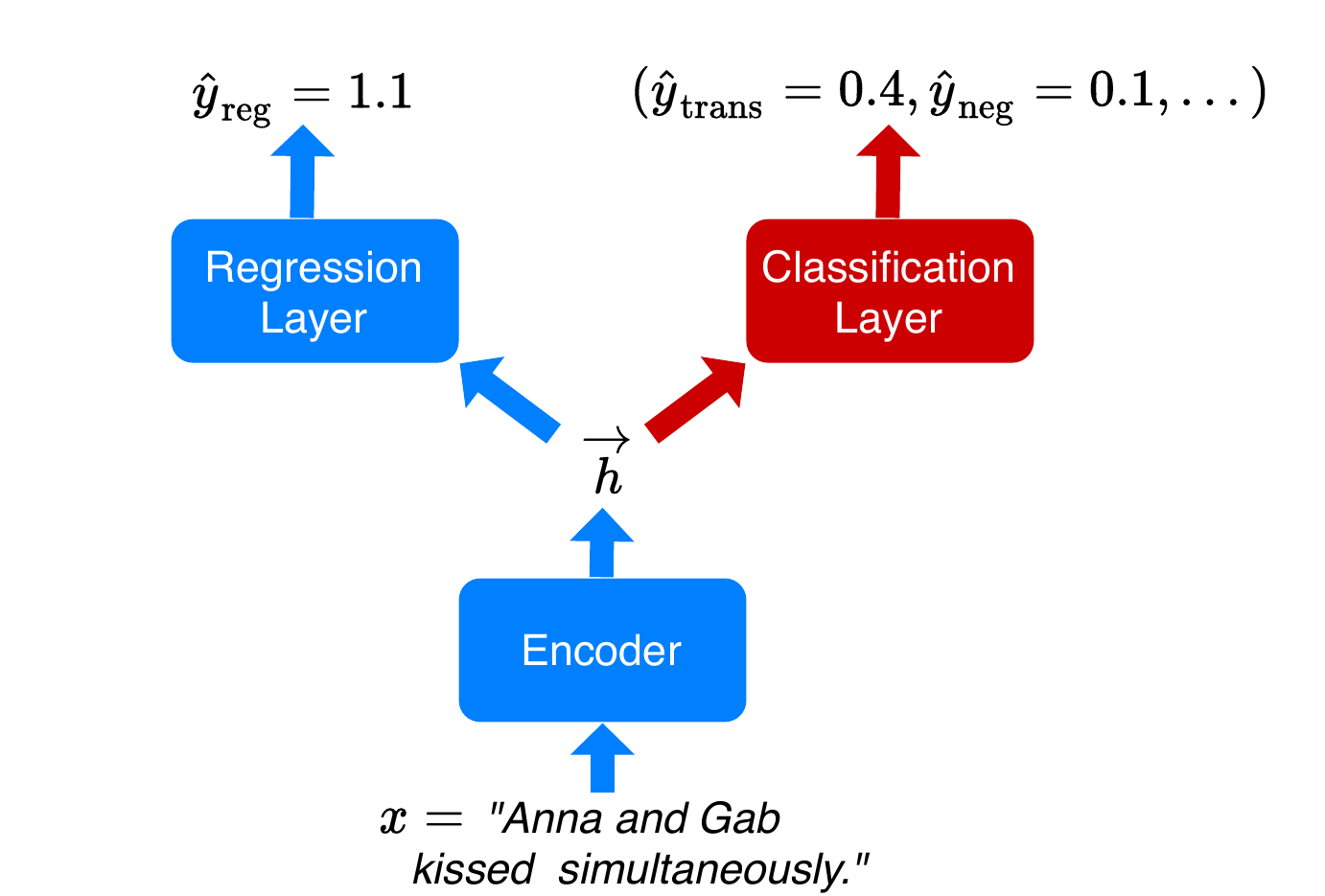}
\caption{Illustration of the methodological framework for symmetry inference. The blue modules represent regression learning pipeline for each single model, and Stage 2 of the hybrid model. The red module denotes Stage 1 of the hybrid model in transfer learning.} \label{framework-diagram}

\end{figure}
\section{Methodology}

We formulate symmetry inference as a regression-based representational learning problem. We explore a set of representational schemes that capture the existing approaches to symmetry based on features and contextualization, as well as a hybrid model that integrates these representations.

As illustrated in Figure~\ref{framework-diagram}, an encoder takes in an input sentence $x^{(i)}$ and provides hidden representation  $\boldsymbol{h}^{(i)} = F(x^{(i)})$ with information pertaining to symmetry inference. An additional regression layer then takes this hidden representation and computes a continuous score that quantifies the degree of symmetry for sentence $x_i$. To adopt a parsimonious approach, we use a simple linear layer $\boldsymbol{w}_\text{reg} \in \mathbf{R}^{dim(\boldsymbol{h})\times1}$ in all experiments, such that 

\begin{equation}
    \hat{y_i} = \boldsymbol{w}_\text{reg}^T \boldsymbol{h}^{(i)} + b_{\text{reg}}
\end{equation}

During the learning process, the model is presented with a ground-truth symmetry score $y^{(i)}$ (from human annotation explained  later) for each sentence. The objective function of the model follows a standard regularized mean-squared loss:
\begin{equation}
    \mathcal{L}_{\text{mse}} = \sum\limits_{i}(y^{(i)} - \hat{y_i})^2  + \frac{\alpha}{2} \boldsymbol{w}_{\text{reg}}^T \boldsymbol{w}_{\text{reg}}
\end{equation}
We keep the regularization hyperparameter $\alpha=1.0$ over all experiments. The parameterized modules of the model are trained to minimize the loss $\mathcal{L}_{\text{mse}}$ via back-propagation. To examine the importance of feature-based and contextual information, we consider four types of encoders with varying model architectures.

{\bf Feature model.}  For each input sentence, the feature-based encoder first  performs dependency parsing, and then extracts a sequence of syntactically-induced, categorical feature variables $\boldsymbol{h}^{(i)}_j$ indicating the existence of certain linguistic patterns. We choose features that were 1) shown empirically to be associated with sentence-level symmetry according to psycholinguistic literature; and 2) obtainable via an automatic feature-extraction pipeline. Following classic empirical studies of symmetry \cite{Gleitman1996SimilarAS}, our model will infer symmetry from pre-defined linguistic features (described in Section~\ref{data}) and a small amount of contextual information from these features (e.g., animacy).

{\bf Static word embedding model.} As a baseline to the contextualized language models, we consider two static embedding encoders, Word2Vec and GloVe \cite{mikolov2013distributed, pennington2014glove}, based on pre-trained distributed word embeddings $\boldsymbol{h}^{(i)}_j$ for each token in $x^{(i)}$, and we then compute the mean vector as  hidden representation:
\begin{align}
    \boldsymbol{h}^{(i)} = \frac{1}{|x^{(i)}|}\sum \limits_{j=0}^{|x^{(i)}|} \boldsymbol{h}^{(i)}_j
\end{align}
As static word embeddings have been shown good at encoding rich lexical knowledge \cite{pennington2014glove}, we expect the simple average representation should capture useful semantic cues beyond the scope of the feature model. 

{\bf Deep contextualized model.} 
To better model  variation in naturalistic context, we use the BERT transformer encoder \cite{devlin2018bert} to compute a contextualized  representation for each input sentence. In particular, we follow the standard approach of applying BERT by considering a special classification token [CLS] at the beginning, and a separation token [SEP] at the end. An encoded vector $\boldsymbol{h}^{(i)}_j$ is then computed for each token, and the last hidden representation of [SEP] is taken as the sentence embedding $\boldsymbol{h}^{(i)}$. As each $\boldsymbol{h}^{(i)}_j$ is a nonlinear function of all input tokens, we expect the contextualized encoders to be superior than static embedding models in extracting more context-sensitive information from  input. The resulting representation is fed into the regression layer to compute a symmetry score. During training, the parameters of the BERT encoder can be either fixed or updated, and we tested in both settings for a comprehensive evaluation.

{\bf Hybrid transfer learning model.}
The richness of knowledge encoded in contextualized models is helpful for many inference tasks, but it might not contain information directly pertaining to predicate symmetry (as made explicit in the feature model). We therefore propose a two-stage transfer learning model to coerce the contextualized model to attend to the symmetry-relevant features, as illustrated in Figure \ref{framework-diagram}. In Stage 1, the BERT encoder is connected with a classification layer with weights $\boldsymbol{W}_{\text{clf}}$, and trained to predict the linguistic features by minimizing the negative log-likelihood loss:
\begin{align}
    \mathcal{L}_{\text{clf}} &= -\sum\limits_{i}\sum\limits_{k=1}^{K} y^{(i)}_k\log(\hat{y}^{(i)}_k) \\
    \hat{y}^{(i)} &= \sigma(\boldsymbol{W}_{\text{clf}}\boldsymbol{h}^{(i)}_j + \boldsymbol{b}_{\text{clf}})
\end{align}
Here $K$ denotes the total number of features for prediction, and $\sigma(\cdot)$ the sigmoid function.  After convergence or in Stage 2, the feature-informed encoder is then topped by a regression layer to produce symmetry scores. This approach incorporates featural knowledge into the existing contextualized model from Stage 1 and transferably applies that knowledge to inferring symmetry in Stage 2.

\section{Symmetry inference sentence dataset  (SIS)} \label{data}

We collect data in two steps: (1) select seed verbs that are traditionally defined as (a)symmetrical and sentences that contain these verbs, and (2) obtain human symmetry ratings for each sentence based on its perceived similarity with its reversed counterpart (e.g., switched  entities) in an online survey.

{\bf Seed verbs.} We focused on verbs because they are the most extensively studied word class in symmetry and have many established features. We worked with 40 common verbs from the literature, divided equally into symmetric and asymmetric categories. Table~\ref{table1} shows the list of verbs. 22 of these verbs are taken from \citet{Gleitman1996SimilarAS}'s original experiments and have thus been previously categorized. The remaining verbs are taken from their reciprocal implication in the Collins English dictionary (\citeyear{collinsdictionary}) and in related literature \cite{Winter2018SymmetricPA, Siloni2012ReciprocalVA}. The selected verbs represent the broad spectrum of symmetry-asymmetry. 

\begin{table}[ht] 
\small
\centering 
    \begin{tabular}{|p{0.1\textwidth}|p{0.32\textwidth}|} 
    \hline 
    {\bf Group} & {\bf Verb predicate} \\
    \hline
    \hline
    Symmetric (20 verbs) & {\it marry,  match, resemble, meet, argue, differ, combine, compare, rhyme, tie, chat, alternate, mix, coexist, clash, converse, collaborate, communicate, agree, separate} \\
    \hline
    Asymmetric (20 verbs) & {\it love, drown, see, hit, follow, choke, eat, copy, save, hate, kill, chase, hurt, push, bounce, break, lecture, hurry, applaud, know}  \\
    \hline 
    \end{tabular}
    \caption{List of verb predicates analyzed in this study.} \label{table1}
\end{table} 

{\bf Sentence extraction.} We semi-randomly extracted 400 sentences (10 sentences per verb) from the English Web 2015 Corpus \cite{Jakubcek2013TheTC} using SketchEngine, a RegEx extraction tool. The chosen sentences contain at least two entities and a verb that denotes some relation between the entities. This relation is structured either as directional or non-directional, with the dataset containing a balanced ratio between the two structures. For the online survey, the sentences are presented with its reversed counterpart, wherein the order of entities is switched. The design of this dataset is based on that of \citet{Gleitman1996SimilarAS}. However, our sentences are different such that their structural and event characteristics naturally vary, while Gleitman and colleagues' test sets strictly contain 2-3 entities and a predicate.

{\bf Feature coding for SIS.}
Table~\ref{table2} summarizes the full set of features used for modeling, using the sentence \textit{I pushed my friends and they pushed me too} as an example. The majority of features have been taken from the literature. Additional features have been selected based on the natural variation of our sentence data, which has not not addressed in previous studies. Most features apply to the clause that contains the entities and target verb predicate, but some account for additional contextual information. We primarily use SpaCy, an open-source library for NLP in Python, for feature extraction. We also use ClausiePy, a SpaCy-based model for clause-based open information extraction \cite{Corro2013ClausIECO}, to extract event-related features (number of entities and number of events). We use the Stanford Named Entity Recognizer with 3-class labels \cite{Manning2014TheSC} to obtain animacy ratings for nouns and pronouns in sentences. One annotator manually corrected the tags assigned by the NER classifier, which were then verified by two more annotators for 10\% of the data. 
We obtained animacy ratings for nouns and pronouns that operated as subject and object in the sentence. A noun is considered animate if their tag is PERSON or ORGANIZATION \cite{Comrie1981LanguageUA}. Animals are manually encoded as animate. Pronouns are considered animate unless explicitly co-referenced with an inanimate entity (e.g., \textit{the walls, they talked to me}). Between annotators, the Kappa statistic \cite{McHugh2012InterraterRT} for the task of rating subject animacy is $\kappa$ = 0.88 whereas for the task of rating animacy matching $\kappa$ = 0.75. Between averaged annotator results and machine ratings, subject animacy is $\kappa$ = 0.75, whereas for animacy matching $\kappa$ = 0.63. Finally, we use the Google Ngram API \cite{Michel176} \footnote{\url{http://storage.googleapis.com/books/ngrams/books/datasetsv2.html}. We determine the prototypicality of the subject entity by extracting and and summing frequencies from 1800 to 2011.}. ``subject + verb'' and ``object + verb'' are represented as strings and later inputted into Google Ngram to obtain their frequencies. If the ``subject + verb'' combination is more frequent, determined by a greater summed proportion, the subject entity is considered more prototypical. If the frequencies are the same, the subject entity is not considered more prototypical.

\begin{table}[ht]
\small
\centering 
    \begin{tabular}{|p{0.39\textwidth}|p{0.04\textwidth}|} 
    \hline 
    {\bf Feature} & {\bf Value} \\ 
    \hline
    \hline
    Is the verb transitive? ($trans$) & 1 \\
    \hline
    Is the verb modified by a preposition? (chat with, $trans\_mod$) &  0 \\
    \hline 
    Is the verb in present tense*? ($v\_tense$) & 0  \\
    \hline
    Is the verb active? ($v\_act$) & 1  \\
    \hline
    Is the verb preceded by a modal expression of uncertainty*? (could, can, might, $modal$) & 0  \\
    \hline
    Is the verb negated*? ($neg$) & 0  \\
    \hline
    Is the verb the root of the sentence*? ($is\_root$) & 1  \\
    \hline
    Is the sentence directional? ($direction$) & 1  \\
    \hline
    Is the entity in subject position singular? ($sing\_sub$) & 1  \\
    \hline
    Is the entity in object position singular? ($sing\_obj$) & 0  \\
    \hline
    Is the entity in subject position conjoined? (\textit{A and B meet}, $conj\_sub$) & 0  \\
    \hline
    Is the entity in object position conjoined? ($conj\_obj$) & 0  \\
    \hline 
    Does the sentence contain a reciprocal phrase? (each other, one another, $rcp\_phrase$) & 0  \\
    \hline 
    Is the subject animate? ($ani\_sub$) & 1  \\
    \hline
    Do the subject and object share the same animacy rating? ($ani\_match$) & 1  \\
    \hline
    Is the subject more frequently paired with this predicate compared to the object? ($sub\_more\_freq$) & 1  \\
    \hline
    How many nominals are in the sentence? ($num\_np$) & 4  \\
    \hline
    How many events are described are in the sentence? ($num\_clauses$) & 2  \\
    \hline
    \end{tabular}
    \caption{Features for symmetry, with example values for \textit{I pushed my friends and they pushed me too.} * denotes new feature not discussed in the literature in relation to symmetry.}  \label{table2}
\end{table}

{\bf Online survey.}
We replicate Experiments 3-4 in \citet{Gleitman1996SimilarAS} study by collecting symmetry ratings with Amazon Mechanical Turk. To ensure the quality of the data, we first ask all online participants to  answer a set of qualification test questions to assess that only native English speakers contribute to the data. Our instructions describe symmetry in sentences as participants  simultaneously being on the giving and receiving end of the action described. Several examples of symmetric and asymmetric sentences are presented. Participants are then presented with pairs of sentences (original and reversed) and asked to rate how similar in meaning the given two sentences are from a scale of 1-5, where 1 indicates the sentences have the same meaning and 5 indicates they do not have the same meaning. Figure ~\ref{fig:sis_amt} shows the instructions provided to the workers. 7 ratings were collected for each of the 400 sentence pairs from 61 workers in total. 

{\bf Data and code availability.} The SIS dataset and code implementation of our symmetry inference methods are publicly available at \url{https://github.com/jadeleiyu/symmetry_inference}.

\begin{figure}[t]
    \centering
    \begin{center}
    \begin{tabular}{|p{0.45\textwidth}|}
    \hline\\
    A sentence is symmetrical if all participants are simultaneously on the giving and receiving end of the action described. If you switch the position of the participants, the overall meaning of the sentence won’t change. \\
    
    In this task, you will be given a pair of sentences. The first describes at least two participants and describes their relationship. The second sentence conveys the same information as the first, except the positions of the participants in the sentence are switched. \\
    
    Your task is to rate how alike in meaning the given two sentences are from a scale of 1-5, where 1 means the sentences do mean the same and 5 means do not mean the same. \\
    
    Given the following pair of sentences: \\
    \begin{enumerate}[label=(\alph*)]

        \item A kisses B on the cheek. 
        \item B kisses A on the cheek.

    \end{enumerate}
       
    Rate how alike in meaning the given two sentences are from a scale of 1-5, where 1 means the sentences do mean the same, and 5 means the sentences do not mean the same.

    \\\hline
    \end{tabular} 
\end{center}
    \caption{Instructions for SIS online survey.}
    \label{fig:sis_amt}
\end{figure}

\section{Model evaluation and results}

We report findings in four steps. First, as a replication we assess the correlation between SIS dataset sentence ratings and verb-level symmetry scores reported  by  \citet{Gleitman1996SimilarAS}.  Second, we evaluate model predictions for sentence-level similarity ratings. Third, we perform an error analysis and interpret the findings. Fourth, we offer a focused analysis of model systematicity.

\subsection{Replicating verb symmetry in SIS dataset}

We average ratings for 10 sentences per verb to represent verb-level scores in the SIS dataset. As the ratings describe similarity in construal, where the lowest rating indicated the highest degree of symmetry, we take the inverse of the average rating to represent verb-level symmetry. For example, if the SIS average similarity rating was 1, its corresponding verb-level symmetry score would be 5. We correlate the resulting 22 SIS verb-level symmetry scores with the corresponding  \citet{Gleitman1996SimilarAS} verb-level symmetry scores and obtain a Pearson's correlation of 0.83 ($p < 0.001$). This finding suggests that the SIS dataset was able to replicate empirical findings at the verb level. We next go beyond the verb-level analysis to evaluate model performance in  predicting symmetry for the naturalistic sentences in the SIS data.




\subsection{Model predictive performance}


To evaluate the models in sentence-level symmetry prediction, we apply a leave-one-predicate-out procedure. Specifically, at each round, sentences associated with one verb predicate are held out as test set, and sentences associated with the remaining 39 verbs are used for training. The hybrid model in Stage 1 is also only trained with features from sentences that do not contain the target verb. The leave-one-predicate-out procedure is repeated 40 times, yielding a predicted symmetry score for every sentence in the SIS dataset. We then correlate the model-predicted symmetry scores against the averaged empirical symmetry ratings from the online survey. We also use mean squared error (MSE), the standard evaluation metric for regression, to evaluate model performance. For all trainings that involves BERT, we use PyTorch-based HuggingFace transformer library to initialize pre-trained BERT encoders. Parameters are updated via Adam optimizer \cite{kingma2014adam}, with learning rate $r=10^{-4}$ and a batch size of 32. 

Table~\ref{table3} summarizes the predictive performance of each model. The static embedding baseline models offer the worst (though statistically significant) prediction, substantially worse than that by the feature model in both Pearson correlation and MSE. We applied a permutation feature importance test \cite{btq134} to the feature model to identify the most predictive features. 7/18 features held positive weight, indicating their usefulness in predicting symmetry. These features were, in order of importance: $conj\_sub$ (0.41), $num\_np$ (0.04), $rcp\_phrase$ (0.03), $sing\_obj$ (0.02), $ani\_match$ (0.02), $direction$ (0.01), and $sing\_sub$ (.005) (see Table~\ref{table2} for description of these features). The contextualized (BERT) model with fine tuning outperforms the feature model in terms of correlation and MSE.\footnote{We also train a model based on BERT encoder without fine-tuning, and obtain MSE = 2.11 and a statistically significant correlation  of 0.49, $p<0.001$.} However, as we show later, despite high overall performance, the contextualized model does not subsume the feature model, and it sometimes erroneously generalizes to unseen test data. The hybrid model offers the best overall performance among all of the models considered, with a near-ceiling correlation and minimal MSE. These results indicate the effectiveness of a joint approach to symmetry inference that combines features with contextual knowledge, and we next interpret and assess errors for the 3 non-baseline models.

\begin{table}[ht]  
\centering 
    \begin{tabular}{|p{0.2\textwidth}|p{0.12\textwidth}|p{0.07\textwidth}|} 
    \hline 
    {\bf Model} & {\bf Correlation}  & {\bf MSE} \\
    \hline
    \hline
    Feature & 0.66 & 1.15  \\
    \hline
    Static Word2Vec  & 0.32 & 1.87 \\
    \hline
    Static  GloVe & 0.33 & 1.85  \\
    \hline
   Contextualized  & 0.79 & 0.87  \\
    \hline 
    \textbf{Hybrid } & \textbf{0.90} & \textbf{0.37} \\
    \hline 
    \end{tabular}
    \caption{Correlations between model predictions and human ratings along with MSE errors. All correlations are significant with $\textit{p} < 0.001$.} \label{table3}
\end{table}

\subsection{Error analysis and model interpretation}



\begin{figure}[h]
\centering 
  \includegraphics[width=1.0\linewidth]{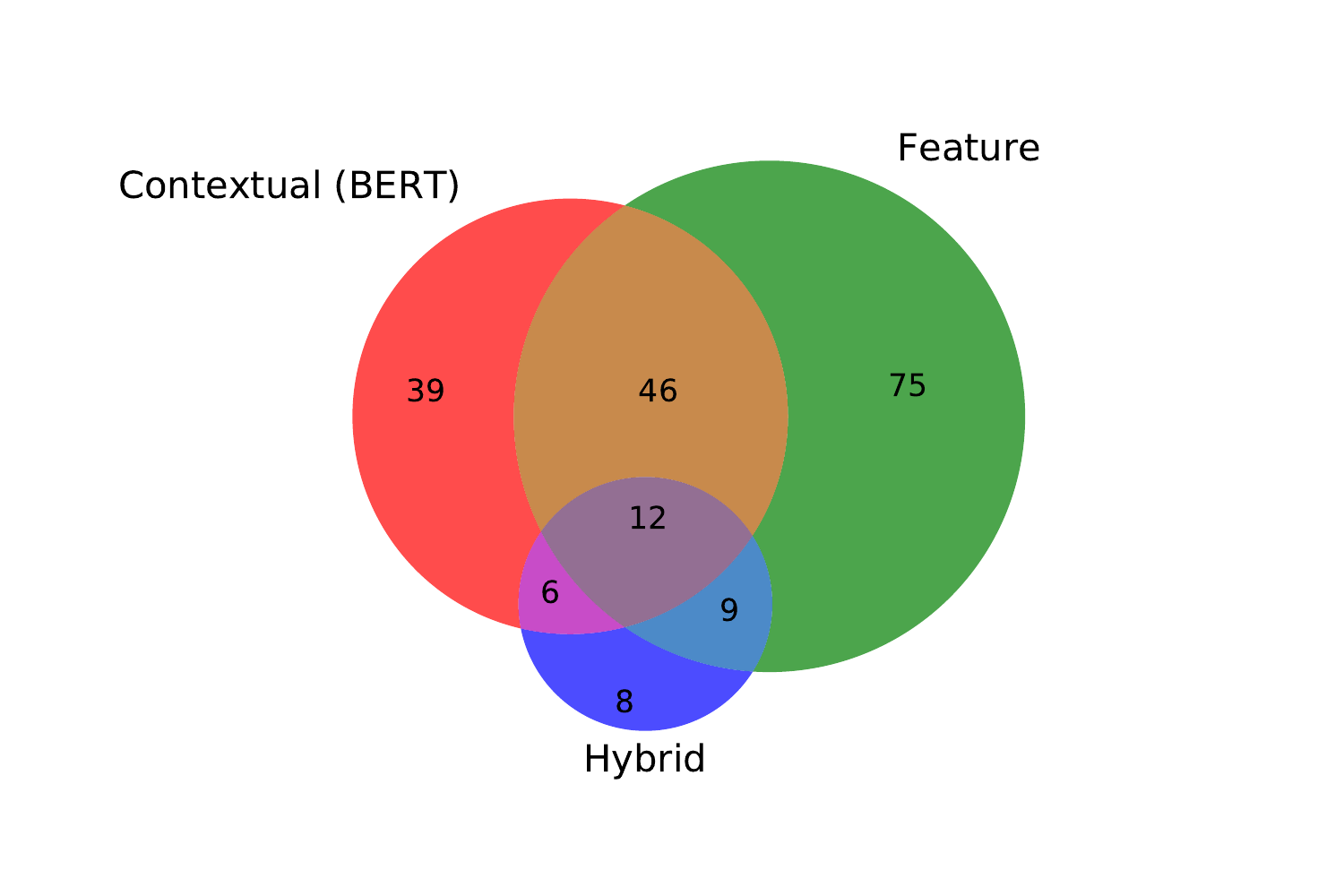}
  \caption{A Venn diagram of number of error cases committed by the three models: feature, contextual (BERT), and hybrid.}
  \label{fig:venn_fig}
\end{figure}

We define error cases as sentences where the absolute difference between the prediction score and the corresponding average empirical rating exceeds 1. Figure~\ref{fig:venn_fig} shows the breakdown of errors committed by the top 3 models: feature, contextualized, and hybrid. The results demonstrate that neither the feature (142/400) nor contextualized (103/400) model can capture symmetry inference alone, reflected in the higher numbers of error cases in comparison to the minimal errors from the hybrid model (35/400). The Venn diagram indicates that a substantial proportion of errors are uniquely committed by the feature model (52.8$\%$) and the contextualized model (37.9$\%$) separately, confirming that these two approaches to symmetry inference contain complementary information. Table~\ref{errors} shows example sentences from model misprediction.

\begin{table*}[ht] 
\small
\centering 
    \begin{tabular}{|p{0.12\textwidth}|p{.82\textwidth}|} 
    \hline 
    {\bf Model } & {\bf Example  sentence of  misprediction} ({\color{red}target verb}, {\color{blue} entities}); ``$\sim$S/AS'' stand for near symmetry/asymmetry\\ 
    \hline
    \hline 
    Feature & 1. \color{blue}This imagery \color{black}will best \color{red}meet \color{blue}your Web needs. {\color{black}(True: $\sim$AS; Human: 4.0; FT Predict: 2.7) } \\
    (FT) &2. \color{blue}I’d \color{black}rather \color{red}collaborate \color{black}with \color{blue}a tarantula \color{black}, because I'm vermin right now. {\color{black}(True: $\sim$AS; Human: 4.3; FT: 2.2)}\\  
    \hline 
     Contextual & 1. \color{blue}The king \color{red}argued \color{black}with \color{blue}the queen \color{black}, scaring the royal advisors. {\color{black}(True: $\sim$S; Human: 2.3; BERT Predict: 3.4)}\\ 
     (BERT) & 2. \color{blue}Some of the nuns \color{black}and \color{blue}the girls \color{black}could \color{red}converse \color{black}fluently in Latin in the Strassburg monastery of St. Margaret. {\color{black}(True: $\sim$S; Human: 1.3; BERT: 2.6)} \\ 
    \hline 
    Shared & 1. \color{blue}I \color{red}applauded \color{blue}my best friend \color{black}and \color{blue}she \color{red}applauded \color{blue}me \color{black}too. {\color{black}(True: $\sim$S; Human: 1.3; FT: 3.1; BERT: 4.1)} \\
    (FT+BERT) & 2. \color{blue}Al-Anon \color{black}can \color{red}hate \color{blue}the alcoholic \color{black}and vice versa. {\color{black}(True: $\sim$S; Human: 1.4; FT: 3.2; BERT: 3.5)} \\
    \hline 
    Hybrid & 1. \color{blue}The doctor \color{red}sees \color{blue}many patients \color{black}, so wait for your turn. {\color{black}(True: S/AS; Human: 2.9; Hybrid: 4.5)} \\
    & 2. \color{blue}The ten-year-olds \color{red}resemble \color{blue}catatonic ghosts plugged into iPods. {\color{black}(True: S/AS; Human: 3.1; Hybrid: 1.5)} \\ \hline
    \end{tabular}
    \caption{Example sentences of predicate symmetry inference where errors were committed by only the feature model, only the contextualized model, shared by both feature and contextualized models, and by the hybrid model, along with suggested  (a)symmetry label of verb-in-context and model prediction (1: symmetric; 5: asymmetric).} \label{errors}
\end{table*} 
  
   The feature model's unique error cases reflect classic problems with the feature-based view of symmetry inference. In the first error case, \textit{imagery} and \textit{web needs} share conceptual similarities and do not exhibit discrepancies in animacy. However, the former is prototypically associated with being an instigator of meeting compared to the latter, which is usually met; this explains why human interpretation was closer to asymmetric. This observation reinforces \citet{Tversky1978StudiesOS}'s discourse on the importance of nominal entity relations. The second error case reiterates the importance of real-world knowledge in addition to understanding Figure-Ground relations \cite{Talmy1985LexicalisationPS}, in that reverse interpretation makes the sentence less natural due to how semantic roles are organized.


The contextualized model's unique errors reflect a possible consequence of having refined knowledge of entities and their real-world relations, such that surface linguistic cues are ignored. The first sample error case is rated symmetrically by human annotators, justified by the conceptual similarity between \textit{king} and \textit{queen}. However, additional contextual information (scaring royal advisors) suggests an influence of historical knowledge of social and gender roles. \citet{Kurita2019MeasuringBI} found that BERT would more strongly associate negative attributes that are especially connotative of authority and power with men, suggesting an inherent gender bias in contextualized word representations. The feature model shows no such bias (for there is no feature that encodes gender or social role). Alternative interpretation is also apparent in the second error case, but at the level of the verb \textit{converse}, which either denotes a symmetrical act of communication or ability to speak in another language. The latter interpretation reduces the symmetrical implication of \textit{converse}, as the relation is no longer reciprocal.

Shared model error cases indicate reciprocity in an additional event, but make no clear attempt to indicate simultaneity. This lack of clarity could justify the discrepancy between contextualized model predictions and human ratings. The feature model cannot infer temporal relations beyond counting the number of events, so their asymmetrical inference was expected. For the second error case in this category, the contextual model's failure may be attributed to knowledge relations or asymmetries in prototypicality between the subject and object, where it is unlikely that Al-Anon, a mutual aid fellowship for alcoholics, would hate an alcoholic.

The hybrid model ratings are more comparable to human annotation, reconciling many of the unique error cases committed by the previous two models. For example, the first feature model error case is reconciled owing to a heightened focus on entity relations. In addition, a larger emphasis on the lexical and animate properties of the entities reconciles the first contextualized model error case. Surprisingly, the model reconciles the first mutually erroneous case, suggesting that some intuition of reciprocity and/or simultaneity has been gained through a stronger consideration and integration of structural features and event characteristics.

The hybrid model's small amount of errors elucidate the consequences of combining feature-based and contextual cues. For example, in the first error case, the predicate \textit{see} might be interpreted symmetrically, given a doctor seeing a patient implies the act of meeting. The human annotators appear to share this sentiment as their similarity ratings are closer to symmetric. However, the directionality of the sentence (linguistic feature) combined with the skewed prototypicality between \textit{doctor} and \textit{patient} with respect to who performs the action of seeing (usually, the patient sees the doctor; contextual feature) invites asymmetrical interpretation. This reasoning can also apply to the second error case. In sum, the hybrid model may reconcile conflicts between using surface linguistic features and context to infer symmetry.

\begin{figure}[t]
\centering 
  \includegraphics[width=1.0\linewidth]{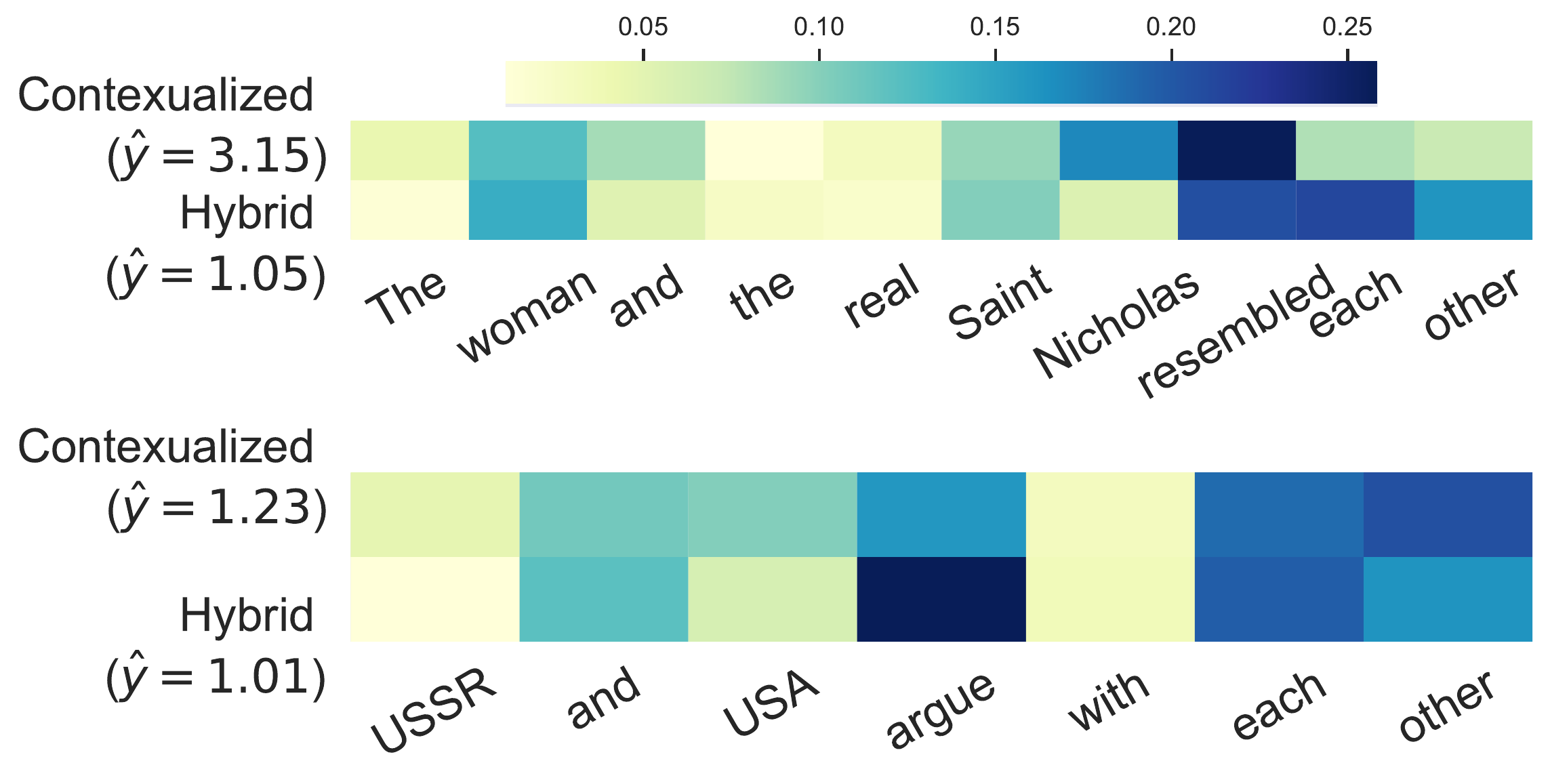}
  \caption{Attention weights and predicted symmetry scores ($\hat{y}$) for contextualized and hybrid models on predicting two symmetric sentences, with target predicates {\it resemble} and {\it argue}, with reciprocal phrases (both have true mean judged score of 1.0).}
  \label{fig:attn-weights}
\end{figure}
\subsection{Focused analysis of model systematicity}

\begin{table}[t]  
\centering 
    \begin{tabular}{|p{0.15\textwidth}|p{0.12\textwidth}|p{0.12\textwidth}|} 
    \hline 
    {\bf Model} & {\bf MSE (ctrl)}  & {\bf MSE (raw)} \\
    \hline
    \hline
    Feature & 0.18 & 1.15  \\
    \hline
   Contextualized  & 0.45 & 0.87  \\
    \hline 
    \textbf{Hybrid } & \textbf{0.17} & \textbf{0.37} \\
    \hline 
    \end{tabular}
    \caption{MSEs on the controlled  (ctrl) and raw sets for feature and contextualized models.} \label{table4}
\end{table}  

We show that certain linguistic cues, such as animacy, are predictive of symmetry and can be easily recognized by humans. To better probe whether contextualized models become more sensitive to such systematic variation after learning, we perform a focused analysis on a subset of SIS sentences controlling for these factors: 1) feature sharing: all sentences that share identical values for a certain set of linguistic features; 2) reliable judged symmetry scores, with low inter-subject standard deviation (we use a threshold $\theta = 0.1$, under 10$\%$ of SD over the dataset $1.20$); 3) decent model prediction, where the differences between the predicted scores and human ratings are low (we use threshold $\theta=1$). Under these criteria, we extract 76 sentences from 5 featural groups. Symmetry of the sentences within each group can be systematically explained  by a certain set of distinct linguistic features (because feature values are shared within each group). 

Table~\ref{table4} summarizes the corresponding MSEs under these controlled subsets, compared to those under the raw whole dataset. The contextualized model---though lower in the raw MSE---is inferior to the feature model in these subsets, because the contextualized model was not able to hone in onto the relevant linguistic cues. In contrast, the hybrid model achieves better performance to the feature model in both controlled and raw data, suggesting that it was able to make systematic generalization that aligns with human judgement. 

To provide intuition on systematicity of the models, we compare the contextualized model with the hybrid model on an example pair of sentences under the distinct linguistic cue of reciprocity (``each other'') for symmetry, and we visualize the attention weights from the final layer of the BERT encoders in Figure \ref{fig:attn-weights}. The heatmap shows that the contextualized model fails to attend to the reciprocal phrase consistently in the two cases (i.e., low attention weights on ``each other'' in the first sentence but high weights in the second sentence), resulting in its poorer generalization. In contrast, the hybrid model assigns high attention weights to ``each other'' in both cases and is therefore performing not only better, but also more systematically.


\section{Conclusion}


We present to our knowledge the first formal framework for modelling sentence-level predicate symmetry and demonstrate that automated inference of verb symmetry is possible in natural context. Contributing the  symmetry inference sentence dataset, we show how existing approaches to symmetry, based on linguistic features and contextualization, are by themselves insufficient to explain sentence-level symmetry judgment, but a hybrid approach improves systematic symmetry inference in state-of-the-art language models. Future work may explore symmetry in other word classes (e.g., nouns and adjectives) and  languages other than English.

\section*{Acknowledgments}
We thank Dzmitry Bahdanau for helpful discussion. AX is supported partly by a UofT Entrance Scholarship. YX is funded through a Connaught New Researcher Award, a NSERC Discovery Grant RGPIN-2018-05872, and a SSHRC Insight Grant \#435190272.

\bibliography{anthology,emnlp2020}
\bibliographystyle{acl_natbib}



\end{document}